\documentclass[runninghead]{llncs}
\usepackage[T1]{fontenc}

\usepackage{booktabs}

\usepackage{xcolor} 
\usepackage{hyperref}
\usepackage{graphicx}
\usepackage{array}
\usepackage{amsmath} 
\usepackage{listings}
\usepackage{algorithm}
\usepackage{algpseudocode}
\usepackage{comment}
    
\begin{document}

\title{LLM-DaaS: LLM-driven  Drone-as-a-Service Operations from Text User Requests}
\author{Lillian Wassim, Kamal Mohamed, and Ali Hamdi}
\institute{\textit{Faculty of Computer Science}\\
MSA University, Cairo, Egypt \\
lillian.wassim, kamal.mohamed, ahamdi @msa.edu.eg
}
%
%

%
%

%
\maketitle              
\begin{abstract}
We propose \textbf{LLM-DaaS}, a novel Drone-as-a-Service (DaaS) framework that leverages Large Language Models (LLMs) to transform free-text user requests into structured, actionable DaaS operation tasks. Our approach addresses the key challenge of interpreting and structuring natural language input to automate drone service operations under uncertain conditions. The system is composed of three main components: free-text request processing, structured request generation, and dynamic DaaS selection and composition. First, we fine-tune different LLM models such as Phi-3.5, LLaMA-3.2 7b and Gemma 2b on a dataset of text user requests mapped to structured DaaS requests. Users interact with our model in a free conversational style, discussing package delivery requests, while the fine-tuned LLM extracts DaaS metadata such as delivery time, source and destination locations, and package weight. The DaaS service selection model is designed to select the best available drone capable of delivering the requested package from the delivery point to the nearest optimal destination. Additionally, the DaaS composition model composes a service from a set of the best available drones to deliver the package from the source to the final destination. Second, the system integrates real-time weather data to optimize drone route planning and scheduling, ensuring safe and efficient operations. Simulations demonstrate the system's ability to significantly improve task accuracy, operational efficiency, and establish \textbf{LLM-DaaS} as a robust solution for DaaS operations in uncertain environments.

\end{abstract}

\section{Introduction}
Drone-as-a-Service (DaaS) is fast emerging as the most prominent business model, which deploys autonomous drones in various applications, including package delivery, infrastructure inspection, and agricultural monitoring \cite{seonjin2022drone,swarm_drones,drones_simulation_based_analysis}. Drones have some unique advantages, such as the ability to move efficiently through crowded city streets and to reduce delivery times significantly compared to traditional methods of delivery \cite{vehicle_routing_drone}.

However, DaaS operations face significant challenges, particularly in changing and uncertain environments. For instance, adverse weather conditions can threaten the stability of drone flights, thereby increasing the risk of delivery failure and operational inefficiencies \cite{drone_weather_flight_uncertainty,AI_Driven_CRM_Chatbot}. To ensure reliable drone operations, it becomes imperative for systems to have the capability to adapt in real-time to such uncertainties \cite{hamdi2020uncertainty,drones_energy_planning}.

One of the immediate challenges of DaaS platforms is making sense of user requests, often in unstructured natural language. Customers usually provide the delivery information in free-text form, which results in a mismatch between such inputs and the structured data needed for efficient drone operations \cite{llm_ai_chatbots,LLM_chatbots}.

Bridging this gap is important to automate DaaS tasks and enhance the efficiency of operations. Toward this, we propose \textbf{LLM-DaaS}, a new Drone-as-a-Service framework that uses LLMs for converting free-text user requests into structured and actionable drone operation tasks \cite{LLM_benchmarking}.

Our system revolves around three main components. First, we introduce an LLM-driven request processing engine that interprets user text inputs, extracting key information such as delivery time, source, and destination locations, and package weight. Second, we incorporate a dynamic DaaS selection and composition model that matches the optimal drone(s) based on the structured request for seamless and efficient package deliveries \cite{drone_flight_scheduling}.

Lastly, the system includes real-time meteorological data to enhance the planning and scheduling of drone routes in order to increase safety and efficiency under uncertain weather conditions \cite{hamdi2020uncertainty,drone_weather_flight_uncertainty,DIJKSTRA_UAV_PATH}.

This work assesses the performance of the \textbf{LLM-DaaS} framework for unstructured text query-to-structured drone task conversion, coupled with simulated DaaS service that showcases its capability in adapting to on-the-spot weather variations. Key metrics including end-to-end time and distance traveled are discussed and compared.

\section{Literature Review} Use of Drone-as-a-Service DaaS, in commercial and industrial applications, has witnessed extensive research in improving operational efficiency and dealing with uncertainties such as dynamic weather conditions and environments. Therefore, this section is going to review state-of-the-art in drone operations, request processing through Large Language Models LLMs, and adaptive path planning relevant to our study.

Managing such uncertainties, such as weather conditions and battery limitations, becomes a very important challenge for DaaS. Hamdi \cite{hamdi2020uncertainty} introduced uncertainty-aware DaaS composition, enabling drones to adjust their flight paths dynamically according to environmental conditions. Similarly, Jiang et al. \cite{seonjin2022drone} studied drone flight scheduling under uncertainty, mainly on battery management and how the weather conditions of wind and temperature would affect the flight performance. Patel et al. \cite{drone_weather_flight_uncertainty} further explore how uncertain weather impacts drone operations, highlighting the requirement for real-time weather adaptation in path planning. These studies, taken together, show the importance of adaptive algorithms in ensuring safe and reliable drone operations in uncertain environments \cite{swarm_drones}.

Although Data-as-a-Service systems heavily depend on structured data to perform any operation, processing free-text user requests is a very serious challenge. Recent research shows great potential in using Large Language Models (LLMs) in automating customer interaction and management of free-text submissions.

Wright et al. \cite{llm_ai_chatbots} demonstrated that LLMs can produce structured outputs for natural language input from users. Again, Anderson et al. \cite{LLM_chatbots} automate the process of developing task-oriented LLM-based chatbots for a range of domains. Also, Wang et al. \cite{LLM_benchmarking} benchmarked the systems powered by LLMs but focused on metrics for the evaluation of their performance.

Our work extends these ideas to apply LLMs to the processing of DaaS-specific requests, such as extracting key details including delivery time, source, destination location, and payload weight from free-text inputs—enabling seamless automation of request handling, a key gap in current DaaS frameworks \cite{llama_2}. Swarm-based drone systems have been proposed to make the DaaS more efficient by collaboration. Lin et al. \cite{swarm_drones} presented a Swarm-based DaaS framework, enabling multiple drones to cooperate in package delivery tasks. Gupta et al. \cite{pathplanning_neural_astar} researched the integration of neural networks to perform path planning in complicated scenarios, further increasing the operations of drones in uncertain environments. Such an approach works well for complex tasks, but our focus in this work is on the optimization of individual drone operations with LLMs to automate free-text request handling. Moreover, LLMs have been integrated into customer-facing systems, for example, CRM \cite{AI_CRM_Chatbots}, in order to improve customer communication during the delivery of service. We extend this use of LLMs to apply them to every step in transforming free-text requests into structured drone tasks in order to improve operational efficiency and customer satisfaction \cite{drone_oneM2M}.
\begin{figure*}[!ht] \centering \includegraphics[width=0.99\textwidth]{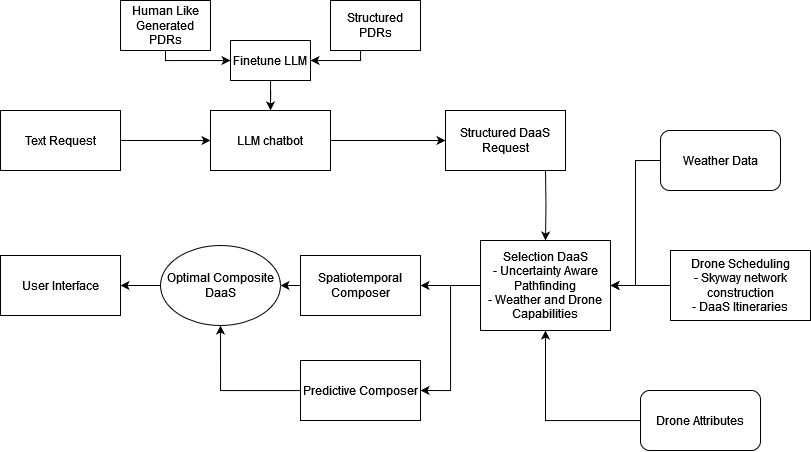} \caption{LLM-DaaS System Architecture} \label{fig
} \end{figure*}

\section{DaaS Problem Formulation}
\subsection{DaaS Service Model}
The DaaS service model is defined as
\[
DaaS = \langle W, P, I \rangle
\]
where:
\begin{itemize}
    \item \( W \) is a set of real-time weather conditions:
    \[
    W = \langle T, WS, WD, H, P \rangle
    \]
    with:
    \begin{itemize}
        \item \( T \) - Temperature at a specific location,
        \item \( WS \) - Wind Speed,
        \item \( WD \) - Wind Direction,
        \item \( H \) - Humidity,
        \item \( P \) - Precipitation.
    \end{itemize}
    \item \( P \) is a set of drone flight paths:
    \[
    P = \langle N, E \rangle
    \]
    with:
    \begin{itemize}
        \item \( N \) - A set of nodes (stations) in the skyway network,
        \item \( E \) - A set of edges representing possible paths between nodes with distances.
    \end{itemize}
    \item \( I \) is the customer interaction log consisting of a set of queries and responses between users and the system.
\end{itemize}
The pathfinding algorithm \( PA \) is a tuple:
\[
PA = \langle H, A^*, D, W_{adj} \rangle
\]
where:
\begin{itemize}
    \item \( H \) is the heuristic function, estimating the distance between two nodes based on their coordinates using the Euclidean distance:
    \[
    H(n, g) = \sqrt{(x_n - x_g)^2 + (y_n - y_g)^2}
    \]
    \item \( A^* \) and \( D \) are the A* and Dijkstra algorithms respectively, both used to calculate the shortest paths between nodes in the skyway network.
    \item \( W_{adj} \) is the weather-adjusted speed, where the drone’s speed \( v \) is adjusted for weather conditions (wind speed \( WS \) and wind direction \( WD \)):
    \[
    v_{adj} = f(v, WS, WD)
    \]
\end{itemize}

\begin{figure}
    \centering
    \includegraphics[width=0.5\linewidth]{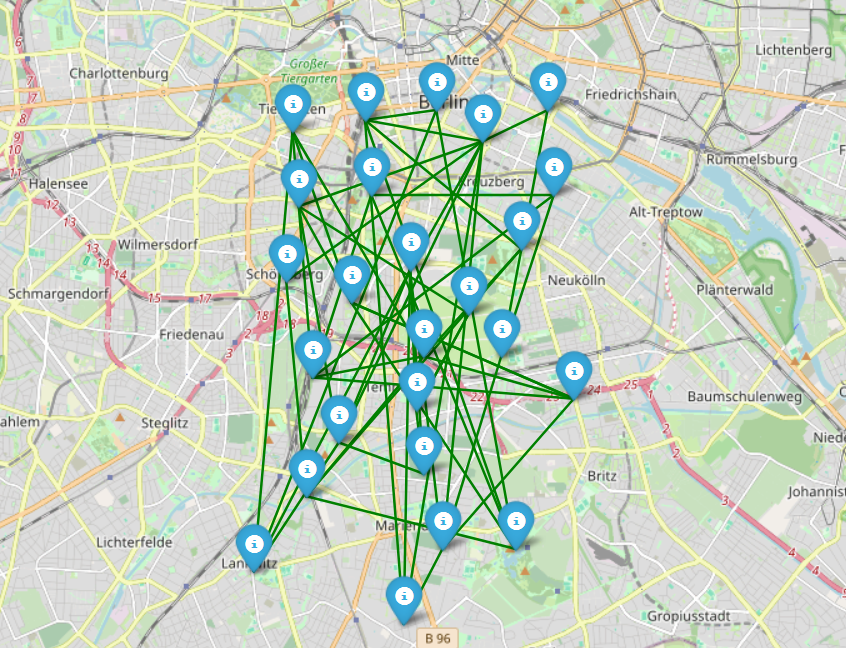}
    \caption{Skyway Map}
    \label{fig:skyway-map}
\end{figure}

\subsection{Real-Time weather adaptation}
Real-time weather data forms the basis of the DaaS system. The system enables the efficient and safe operation of drones in the face of very dynamic and unpredictable conditions. Weather plays a large role in the performance and safety of flights; thus, the system must be able to make quick adaptations to changing conditions.The system uses meteorological data, which is stored in CSV files and encompasses much information on temperature, wind speed, humidity, and precipitation for different geographical areas. The weather adaptation module reads this information, which the system uses to modify the paths and speeds of the drones according to the change in conditions. The main components of the adaptation include:
\begin{itemize}
\item Dynamic Route Adjustment: The system is supposed to monitor the main meteorological parameters continuously, such as wind speed and direction, temperature, humidity, and precipitation. When the detection of bad weather happens, pathfinding algorithms like A* or Dijkstra's algorithm are used to calculate an alternative route for the drone to fly around dangerous areas or less desirable paths.
\item Velocity Variations: Meteorological conditions can quite drastically affect the velocity and stability of an unmanned aerial vehicle. For instance, strong winds may decelerate a drone or even break the stability of its flight. The system adjusts the velocity of the drone depending on the weather conditions to efficiently ensure safe navigation.

\end{itemize} 
\subsection{Drone Fleet and Skyway Network} The drone fleet is operated within a controlled aerial transportation system whereby a network of delivery stations are linked to each other by predefined routes. Each drone is assigned to specific stations, using both A* and Dijkstra's algorithms to optimize routing, thus ensuring that the fastest and safest routes are chosen. The network contains many delivery locations, and throughout every delivery mission, drones fly through multiple nodes.
This includes maintenance schedules and battery recharging, providing detailed timelines for drones based on their usage and operational efficiency.

\subsection{Drone Attributes}
Simulation: Each drone will have a set of attributes that affect its performance, including:

\begin{itemize}
\item Battery Health: Simulated battery degradation over time affects flight range and recharging frequency.

\item Maintenance Status: Drones need periodic maintenance based on flight hours and operational stress, simulated to ensure realistic downtime.

\item The payload capacity is not the same for various unmanned aerial vehicles; the system simulates delivery situations that use different weights of payloads. 
\end{itemize}
\subsection{Simulation and Data Acquisition}

Moreover, this simulation represents a wide range of environmental factors, including dynamic weather conditions like wind speed, temperature, and precipitation—that all impact the effectiveness of drone operations and their decision-making related to navigation. The model includes realistic geographical constraints for different regions; every station or waypoint has specific geographic coordinates assigned to it, hence allowing exact control over route optimization by following airspace constraints and regional legislation.

All simulated flights are stored in individual flight logs, which include details of each trip segment, including data such as distance traveled, flight duration, remaining battery capacity, and maintenance status. Such extensive data collection ensures that all aspects of drone operation are monitored, thus allowing assessment of performance under different environmental conditions.

Additionally, when there are complications like route disturbances or maintenance that needs to be carried out, these are all recorded along with corresponding error messages, which are then used to simulate and test error handling within the DaaS system. Long-term system performance is assessed by dividing the simulation into monthly-sized time slots over a period of four years. Each slot represents unique combinations of meteorological conditions and operational scenarios to simulate seasons, peak-demand periods, and service disruptions. In this way, the simulation will be able to explain trends in resource usage, delivery completion rates, and maintenance needs under more realistic and dynamic conditions. The monthly collected data allows for analyzing over time several parameters like battery degradation, recurrent faults, or optimal flight paths. As such, this simulation data provides a wide base for analyzing the efficiency, reliability, and resilience of DaaS systems under different operating conditions. This long-term data allows for trend analysis and further refinement of drone performance models to continue the optimization of the skyway network for DaaS operations over time.

\section{LLM DaaS} 
\subsection{LLM Finetuning} The LLM component is, in essence, central to the ability of the system to understand free-form, unstructured human input and to convert that into actionable, structured delivery requests. In the proposed system, users submit natural language inputs that contain delivery requirements including pickup location, destination, and package weight. The LLM understands the input and converts it accordingly. The conversion takes place in a multi-step process. First, the LLM analyzes and partitions the input to identify and extract key entities, such as addresses, package information, and time constraints. Using the contextual information and predefined rules, it maps these entities to specific fields needed by the DaaS framework for the operation of a drone in pickup, destination, and payload. Upon completion of text processing, the extracted information is structured in a formalized delivery request format. This format is suitable for the DaaS model, where the drone resources are managed, navigation is taken care of, and task allocation happens. The formalized request is then pushed to the core modules of the system, which can determine the optimal route, assess meteorological conditions, and allocate drones accordingly. Such an approach by LLM would make the system friendlier, allowing for easy interaction with nontechnical users and a backend that could work well because of the well-structured inputs.

\begin{table*}[!h]
\centering
\caption{Examples of structured delivery requests and corresponding free-text user inputs}
\label{tab:request_examples}
\begin{tabular}{@{}p{0.3\textwidth}p{0.65\textwidth}@{}}
\toprule
\textbf{Structured Delivery Request}  & \textbf{Free-Text User Request Example} \\ \midrule
\texttt{request\_id=1, start\_node=7, destination\_node=14, payload=3kg} & "Hi, please help me pick up my package from home at node 7 and deliver it to my workplace at node 14. The box weighs around 3 kilograms. Thanks!" \\ \midrule
\texttt{request\_id=61, start\_node=18, destination\_node=1, payload=7kg} & "Good afternoon! I’d like to request a delivery. The payload weighs 7 kg and the pickup is at node 18. The drop-off location is node 1. Could you confirm the exact drop-off time once it’s done? Also, if possible, I’d like a picture of the package when it’s delivered." \\ \midrule
\texttt{request\_id=77, start\_node=20, destination\_node=14, payload=3kg} & "Hello, I need to schedule a delivery for a 3 kg package. The pickup is at node 20, and it needs to be dropped off at node 14. Could you also check the drone’s battery before sending it out?" \\ 
 \\ \bottomrule
\end{tabular}
\end{table*}

\subsection{Data Collection}
Fine-tuning the LLM models for structured data generation; hence, this paper proposed a two-step process in this regard.
\begin{itemize}
\item \textbf{Structured Delivery Requests}: We generated 5,000 structured delivery requests with mandatory details such as \texttt{request\_id}, \texttt{start\_node}, \texttt{destination\_node}, and \texttt{payload}. The requests form the basis of DaaS operations in that they provide the primary metadata needed for the processing of deliveries. An example of a structured request would look like: \texttt{request\_id=1, start\_node=7, destination\_node=14, payload=3kg}.
\item \textbf{Free-Text Requests via Prompt Engineering:} We employed the structured delivery requests to create natural-sounding free-text requests utilizing ChatGPT. using a prompt engineering technique, we fed the structured requests along with their respective free-text exemplars to create conversational versions of the structured data. Table~\ref{tab:request_examples} shows a number of examples of structured delivery requests and their corresponding free-text versions. This dual-format dataset is used to fine-tune LLMs that can convert real users' input into structured DaaS operations. \end{itemize}
\subsection{Formulating Delivery Request using LLM}

To better understand how LLMs do in translating free-text requests from users into structured operations (DaaS), we evaluate their performance with a few experiments. The complete process is delineated below:

\begin{enumerate}
\item \textbf{Preliminary Test}: The LLMs are first fine-tuned and tested using 1,000 free-text requests. The notion is to check their efficiency in the accurate translation of the requests into structured delivery requests. The main fields targeted in extraction being \texttt{request\_id}, \texttt{start\_node}, \texttt{destination\_node}, and \texttt{payload}.

\item \textbf{Fine-Tuning}: Each LLM is fine-tuned on a training dataset consisting of 4,000 free-text requests along with their corresponding structured outputs. In this step, the models are expected to learn the specific mapping between the user inputs and the structured data necessary to drive DaaS successfully.

\item \textbf{Post-Fine-Tuning Evaluation:} We re-evaluated all LLMs after fine-tuning on the same set of 1,000 test requests to measure any improvement in their performance. This evaluation aimed to quantify the enhancements achieved through the fine-tuning process.

\item \textbf{Evaluation Metric:} The \textbf{G-Eval metric} was used for evaluation, relying on chain-of-thought (CoT) reasoning in LLMs. This allows nuanced assessment of the model outputs against custom criteria, enabling us to not only judge surface-level accuracy but also contextual understanding and logical coherence in converting the user's inputs into structured DaaS operations.
\item \textbf{Benchmark Comparison:} The comparison between pre- and post-fine-tuning was finally carried out so as to quantify the improvement in performance for each of the diverse test cases. The benchmark comparison is indispensable to check the validity of the fine-tuning process and assure very good performances of LLMs for real-world applications in the DaaS framework.

\end{enumerate}
\subsection{Drone Selection and Composition}
The DaaS Selection and Composition aspects are critical in optimizing effective and efficient delivery operations. In that respect, the system has an advanced evaluation process through which it selects for every delivery task the most appropriate drone(s) from the fleet. The system considers a set of parameters to ensure that the selected drone will properly execute the delivery task; these include:
\begin{itemize}

\item Battery Life and Health Status: All drones have an operating time limit depending on the battery capacity of the drone. The system checks the remaining battery life of the assigned drones, ensuring that the drone has more than enough to complete the delivery without an unplanned recharge.
\item Payload Capacity: Drones have different payload capacities that determine their ability to carry packages with regards to size and weight. \item Flight Speed and Range: Drones have varying speeds and ranges of operation. The selection system takes this into consideration to match the drone's capability with the delivery payload.

\end{itemize}

The Composition capability of the system allows it to coordinate multiple drones for more complex delivery tasks or longer routes. For example, when the delivery covers a longer distance or has multiple points for delivery, the system will use drones intelligently to offload or exchange packages at intermediate nodes. The system hence has increased efficiency and reliability of delivery, since it can dynamically reassign packages to more suitable drones in midroute, taking into consideration the state of each drone's battery life, health status, speed, and local weather conditions. With this multi-drone deployment, complex deliveries are handled in a seamless manner, keeping the DaaS network resilient, flexible, and at high service levels—irrespective of external challenges.

\section{Results}

\subsection{LLM Performance Comparison}

This section evaluates several Large Language Models (LLMs) for transforming free-text user requests into structured Drone-as-a-Service (DaaS) tasks, including Gemma 2b, LLaMA 3.2, Phi-3.5, and Qwen-2.5, both before and after fine-tuning. The evaluation utilized the G-Eval score from the DeepEval framework, focusing on the models’ ability to accurately extract delivery locations, times, and package specifications.

The results indicate that fine-tuning significantly improved the models’ performance, with some achieving G-Eval scores close to 95\%, demonstrating their enhanced capability in converting unstructured text into actionable DaaS tasks.

\begin{itemize}
    \item \textbf{Gemma 2b}: Initially performed poorly with a low G-Eval score, struggling with ambiguous inputs. After fine-tuning on a specialized dataset, it showed substantial improvement, becoming capable of accurately structuring requests into DaaS tasks. (See Appendix, Table~\ref{tab:gemma_model_scores}).
    
    \item \textbf{LLaMA 3.2}: Achieved the highest G-Eval score (0.9977), effectively handling diverse and complex user inputs. Its architecture allowed for exceptional precision in interpreting vague requests, making it highly suitable for real-world applications. (See Appendix, Table~\ref{tab:llama_model_scores}).
    
    \item \textbf{Phi-3.5}: Also showed significant post-fine-tuning improvement, achieving a strong G-Eval score (0.9977). Although smaller than LLaMA 3.2, it maintained high accuracy and efficiency in real-time processing, making it a competitive choice for scenarios requiring both speed and precision. (See Appendix, Table~\ref{tab:phi_model_scores}).
    
    \item \textbf{Qwen-2.5}: Despite being the smallest model, it performed well after fine-tuning, achieving a respectable G-Eval score of 0.9887. Initially limited in handling complex queries, it improved in interpreting and structuring simpler requests, proving that smaller models can benefit from targeted fine-tuning. (See Appendix, Table~\ref{tab:qwen_model_scores}).
\end{itemize}

Overall, the fine-tuning process substantially improved the performance of all evaluated models. The G-Eval scores, which reached nearly 99\% for several models, underscore that LLMs can reliably transform free-text user requests into structured DaaS tasks with exceptional precision. \textbf{LLaMA 3.2} emerged as the top-performing model, offering an outstanding balance between handling complex inputs and maintaining high accuracy. \textbf{Gemma 2b} and \textbf{Phi-3.5} demonstrated strong post-fine-tuning performance, while \textbf{Qwen-2.5}, despite its smaller size, showed significant improvements, proving that even compact models can benefit greatly from targeted fine-tuning.

\subsection{Algorithms for Pathfinding}
This section presents a comparative analysis of the \textit{Dijkstra} and \textit{A*} algorithms, conducted under uniform meteorological parameters (specifically wind speed, temperature, and wind direction), emphasizing the duration of flight and the distance traversed by the drones along each pathway.
\subsubsection{Route 46 Comparison}
As for Route 46, both algorithms went through the same weather conditions: wind speed 14.9 m/s, wind direction 135°, temperature 9.5°C. Comparative performance analysis between the two algorithms is given below:
\begin{itemize}
\item \textbf{Dijkstra:} Total flying time was \textbf{1 hour, 57 minutes} over a distance of \textbf{119.08 km}.
\item \textbf{A*} The flight took slightly less time at \textbf{1 hour, 51 minutes}, covering a slightly smaller distance of \textbf{113.91 km}.
\end{itemize}

Although the weather was exactly the same, \textbf{A*} performed better by choosing a shorter path, hence the actual distance flown and the time taken were shorter.

\subsubsection{Route 47 Comparison}
For Route 47, both algorithms again faced the same weather conditions. Their performance comparison is as follows:
\begin{itemize}
\item \textbf{Dijkstra:} The total flying time was \textbf{2 hours, 59 minutes}, for a distance of \textbf{179.08 km}.
\item \textbf{A*:} The flight time increased to \textbf{3 hours and 4 minutes} and the total distance covered was \textbf{187.00 km}.

\end{itemize}

Here, \textbf{Dijkstra} managed to get a shorter flight time and covered a smaller distance, while \textbf{A*} picked a slightly longer path.

\section{Conclusion}

In the present study, a comprehensive Drone-as-a-Service (DaaS) framework is presented to improve operational efficiency in unpredictable scenarios by integrating Large Language Model capabilities with real-time weather adaptation. Our technology ensures secure and timely delivery even in the most challenging conditions, thanks to the optimization of flight operations using advanced pathfinding algorithms coupled with up-to-date meteorological information. By integrating the LLMs, the system is capable of automating the transformation of free-text user requests into structured and actionable drone operation tasks. It saves a lot of manual intervention and improves the overall efficiency of service while enhancing the user experience. Our experimental results indicate that this system could achieve near-to-perfect accuracy in extracting the structured data from user inputs—a clear demonstration of the effectiveness of fine-tuning LLMs for DaaS applications. Future research will focus on further improvement of the pathfinding algorithms and extension of the functionality of the system in order to be able to handle larger fleets and more complex scenarios of delivery. We will also look into the possibility of integrating state-of-the-art machine learning methods to further raise the operational cost-effectiveness and route optimization. By these efforts, we surely believe that our system will definitely emerge as a credible alternative for DaaS operations in uncertain and dynamic environments.

\section{Acknowledgment}
We extend our heartfelt gratitude to AiTech AU, \href{https://aitech.net.au}{\textit{AiTech for Artificial Intelligence and Software Development}}, for funding this research, providing technical support, and enabling its successful completion.

\section{Data and Code Availability}
The dataset generated can be found in the code repository along with the code and subsequent findings \href{https://github.com/KamalM13/graduation-project}{repo}

\bibliographystyle{bibtex/spmpsci}
\bibliography{ref}
\newpage

\appendix
{\textbf{Appendix A: }G-Eval Scores and Explanations for Different Models}
\begin{table}[ht]
\small
\centering
\caption{Gemma 2b G-Eval scores and explainations}
\label{tab:gemma_model_scores}
\begin{tabular}{@{}p{0.1\textwidth}p{0.1\textwidth}p{0.1\textwidth}p{0.7\textwidth}@{}}
\toprule
\textbf{Test Case} & \textbf{Model Type} & \textbf{Score} & \textbf{Explanation} \\ \midrule
Test Case 1 & Pretrained & 0.996 & The actual output correctly reflects the start node 24, destination node 32, and payload 5.01 kg as per the input, without contradictions or vague language, and in structured data format. \\ \midrule
Test Case 1 & Non-Pretrained & 0.31 & Mentions picking up the parcel at node 24, but does not confirm the delivery to node 32 or specify the payload of 5.01 kg. \\ \midrule
Test Case 15 & Pretrained & 0.98 & The actual output correctly reflects the start node as 36, the destination node as 22, and the payload as 4.2 kg, with no contradictions or unstructured data. \\ \midrule
Test Case 15 & Non-Pretrained & 0.254 & The output lacks structured data and does not mention start node 36 or destination node 22, only acknowledges the 4.2 kg payload  \\ \midrule
\end{tabular}
\end{table}

\begin{table}[!h]
\small
\centering
\caption{Qwen 2.5 G-Eval scores and explanations}
\label{tab:qwen_model_scores}
\begin{tabular}{@{}p{0.1\textwidth}p{0.1\textwidth}p{0.1\textwidth}p{0.7\textwidth}@{}}
\toprule
\textbf{Test Case} & \textbf{Model Type} & \textbf{Score} & \textbf{Explanation} \\ \midrule
Test Case 11 & Pretrained & 0.9887 & The actual output correctly identifies the start node as 2, destination node as 31, and payload as 7.48 kg, following the input criteria without contradictions or vague language. \\ \midrule
Test Case 11 & Base & 0.2334 & The output does not mention node 2 or node 31, which are crucial facts from the input. \\ \midrule
Test Case 15 & Pretrained & 0.9673 & The actual output correctly identifies the start node, destination node, and payload as specified in the input. However, it doesn't address checking the drone's battery, which is required before sending it out. \\ \midrule
Test Case 15 & Base & 0.1298 & The actual output lacks structured data and does not mention the start node or destination node as specified in the input. It is incomplete and uses vague language, failing to address the specifics of the delivery details provided. \\ \midrule
\end{tabular}
\end{table}

\begin{table*}[!h]
\centering
\caption{Phi 3.5 G-Eval scores and explanations}
\label{tab:phi_model_scores}
\begin{tabular}{@{}p{0.1\textwidth}p{0.1\textwidth}p{0.1\textwidth}p{0.7\textwidth}@{}}
\toprule
\textbf{Test Case} & \textbf{Model Type} & \textbf{Score} & \textbf{Explanation} \\ \midrule
Test Case 1 & Finetuned & 0.9600 & The actual output contains all required structured data: start node 14, destination node 33, and payload 7.99 kg, without contradictions or vague language. \\ \midrule
Test Case 1 & Base & 0.2979 & The output confirms the need to fully charge the drone, aligning with the input, but fails to specifically mention the start node 14, destination node 33, or the payload of 7.99 kg, and uses non-structured data. \\ \midrule
Test Case 2 & Finetuned & 0.9977 & The actual output correctly includes structured data with start node 2, destination node 36, and payload 3.62, matching the relevant information from the input. \\ \midrule
Test Case 2 & Base & 0.2259 & The actual output does not contradict the facts but fails to mention destination node 36 and does not address payload or alternate routes, which are crucial details. \\ \midrule
\end{tabular}
\end{table*}

\begin{table}[]
\centering
\caption{Llama 3.2 G-Eval scores and explanations}
\label{tab:llama_model_scores}
\begin{tabular}{@{}p{0.1\textwidth}p{0.1\textwidth}p{0.1\textwidth}p{0.7\textwidth}@{}}
\toprule
\textbf{Test Case} & \textbf{Model Type} & \textbf{Score} & \textbf{Reason} \\ \midrule
Test Case 1 & Finetuned & 0.831 & The start and destination node numbers are mentioned as well as payload weight. Language is specific. \\ \midrule
Test Case 1 & Base & 0.32 & The plan doesn't include the start node and destination node. The task details don't match 'pickup at node 14' and 'drop-off at node 33'. The package's weight is provided. \\ \midrule
Test Case 10 & Finetuned & 0.845 & The expected output has a clear request, start node, destination node, and payload. However, the actual output lacks the request. \\ \midrule
Test Case 10 & Base & 0.24 & The text includes specific details about the shipment, such as weight and location, but fails to provide clear and direct instructions. \\ \midrule
\end{tabular}
\end{table}

\end{document}